\documentclass{article}
\usepackage{ijcai22}

\usepackage{times}
\usepackage{soul}
\usepackage{url}
\usepackage[utf8]{inputenc}
\usepackage[small]{caption}
\usepackage{graphicx}
\usepackage{amsmath}
\usepackage{amsthm}
\usepackage{booktabs}
\usepackage{algorithm}
\usepackage{algorithmic}
\usepackage{amsfonts} %
\usepackage[hidelinks]{hyperref}

\usepackage{graphicx}%

\usepackage{stfloats}
\usepackage{colortbl}
\definecolor{mygray}{gray}{.9}

\usepackage{threeparttable}
\usepackage{multirow}
\usepackage{booktabs}

\title{
Entity Alignment with Reliable Path Reasoning and Relation-Aware Heterogeneous Graph Transformer
}

\author{
	Weishan Cai$^{1,2}$\and
	Wenjun Ma$^1$\and
	Jieyu Zhan$^1$\and
	Yuncheng Jiang$^{1,3}$\footnote{Corresponding Author}
\affiliations
$^1$School of Computer Science, South China Normal University, China\\
$^2$School of Computer and Information Engineering, Hanshan Normal University, China\\
$^3$School of Artificial Intelligence, South China Normal University, China
\emails
	caiws@m.scnu.edu.cn,
	phoenixsam@sina.com,
	zhanjieyu,ycjiang@scnu.edu.cn
}

\begin{document}
\maketitle
\begin{abstract}
	Entity Alignment (EA) has attracted widespread attention in both academia and industry, which aims to seek entities with same meanings from different Knowledge Graphs (KGs). 
	There are substantial multi-step relation paths between entities in KGs, indicating the semantic relations of entities. 
	However, existing methods rarely consider path information because not all natural paths facilitate for EA judgment. 
	In this paper, we propose a more effective entity alignment framework, \textit{RPR-RHGT}, which integrates relation and path structure information, as well as the heterogeneous information in KGs. 
	Impressively, an initial reliable path reasoning algorithm is developed to generate the paths favorable for EA task from the relation structures of KGs.
	This is the first algorithm in the literature to successfully use unrestricted path information. 
	In addition, to efficiently capture heterogeneous features in entity neighborhoods, a relation-aware heterogeneous graph transformer is designed to model the relation and path structures of KGs. 
	Extensive experiments on three well-known datasets show \textit{RPR-RHGT} significantly outperforms 10 state-of-the-art methods, exceeding the best performing baseline up to 8.62\% on Hits@1. 
	We also show its better performance than the baselines on different ratios of training set, and harder datasets.
\end{abstract}

\section{Introduction}
\label{sec1}

Most Knowledge Graphs (KGs) are often disconnected from each other because they are constructed with different technologies and languages, which poses challenges for merging and integrating different KGs.
Entity Alignment (EA) is a task to connect entities with the same meaning in different KGs, which plays a fundamental role in the knowledge fusion of KGs.
Recently, EA methods based on the Graph Neural Networks (GNNs) are more favored by researchers than the translation-based methods.
GNNs not only exhibit excellent performance in aggregating the neighborhood features of nodes, but also can design corresponding feature acquisition methods for EA tasks, while translation-based methods are designed for link prediction.

Although current GNNs-based methods have achieved promising results, they still suffer from the following three limitations.
First, many methods \cite{wu_relation-aware_2019,sun_knowledge_2020} treat KGs as homogeneous graphs without considering the heterogeneous features of sides between entities. 
Actually, the heterogeneous information helps to improve the accuracy and robustness of alignment judgments.
Second, some semantic information other than relation structures is considered by many works, 
such as entity attributes \cite{liu_exploring_2020,cai_multi-heterogeneous_2022}, 
text descriptions \cite{yang_aligning_2019}, and multi-modal information \cite{liu_visual_2021}.
However, the more semantic information a method integrates, the more data its application requires, which cannot be satisfied in some scenarios. 
Third, some other works \cite{wu_neighborhood_2020,zhu_collective_2020} only rely on the relation structures, and obtain inter-graph information based on Graph Matching Networks (GMN) \cite{li_graph_2019} to mine more similar features between aligned entities.
Nonetheless, the matching module they introduced for learning inter-graph information runs through the entire training process with high temporal and space complexity.

Therefore, for the first limitation above, we design a \textit{R}elation-Aware \textit{H}eterogeneous \textit{G}raph \textit{T}ransformer (\textit{RHGT}) to effectively extract the similarity features of aligned entities in their heterogeneous structures.
For the latter two limitations above, we develop a \textit{R}eliable \textit{P}ath \textit{R}easoning algorithm (\textit{RPR}) that can directly extract the path structures favorable for EA tasks from the original relation structures.
Existing methods rarely consider the path information of KGs (i.e., the indirect neighborhood of aligned entities), despite their success in modeling of direct relationship facts. 
It is known that substantial multi-step relational paths exist between entities, indicating their semantic relationships.
But not all natural paths facilitate EA judgment, and some even backfire.
Although IPTransE \cite{zhu_iterative_2017} considers the reliability of paths, it assumes all relations between KGs are pre-aligned.
Essentially, our idea is to make full leverage of the richness of KGs by simultaneously comparing the similarities of relation and path structures of aligned entities. 
We believe the paths that frequently appear near pre-aligned entities can be regarded as reliable and used to align other entities.
The fusion of relation and path structure information complements each other, alleviating the inconsistency between each type information of aligned entities.

After all, we combine above two methods into a entity alignment framework called \textit{RPR-RHGT}, which not only considers the heterogeneous information of sides in KGs, but also mines the path information within the relation structures of KGs.
Extensive experiments on three well-known benchmark datasets show \textit{RPR-RHGT} not only outperforms 10 state-of-the-art models significantly, but also has impressive scalability and robustness.

\section{Related Work}
\label{sec2} 
\paragraph{Translation-based Entity Alignment.}
Such methods are mainly based on TransE \cite{bordes_translating_2013} and its variants. %
MTransE \cite{chen_multilingual_2017} is the pioneering work, which uses TransE to model the entities and relations and evaluates the transform between two vector spaces based on pre-aligned entities.
Other works utilize additional information or external knowledge of KGs, such as 
attribute structures \cite{sun_cross-lingual_2017,zhang_multi-view_2019,trisedya_entity_2019}, 
entity descriptions \cite{chen_co-training_2018}, entity names \cite{zhang_multi-view_2019}, 
ontology schemata \cite{xiang_ontoea_2021}, 
to find more similar features of aligned entities.
There are also some works \cite{sun_bootstrapping_2018,zhu_iterative_2017} that try to discover more new aligned entities by iterative strategies.

\paragraph{GNNs-based Entity Alignment.}
GNNs-based methods mainly utilize Graph Convolutional Networks (GCNs) and Graph Attention Networks (GATs) to aggregate the neighborhood feature of each entity, thereby obtaining the neighborhood similarity between aligned entities.
Most of them directly compare the neighborhood similarity between aligned entities in relation structures \cite{wu_relation-aware_2019}.
There are several attemptsto simultaneously consider the similarities in the attribute and relation structures \cite{liu_exploring_2020}. 
Some other works smartly model both intra-graph and cross-graph information, and learn similarities by building cross-graph attention mechanism using GMN \cite{wu_neighborhood_2020,zhu_collective_2020}. %
Besides, some researchers believe that the heterogeneity of edges in KGs should be considered when aggregate the neighborhood features, because KGs are heterogeneous graphs.
They propose or apply some heterogeneous graph embedding methods to learn better representations for entities \cite{cai_multi-heterogeneous_2022}. 
All aforementioned works only consider the similarity of direct neighborhoods between aligned entities.
However, aligned entities have some similarity in their indirect neighborhoods.
Hence, we attempt to obtain the similarity between aligned entities in the relation structures and multi-hop path structures of KGs simultaneously in the paper.

\paragraph{Heterogeneous GNNs.}
Recently, many works have tried to extend GNNs to the modeling of heterogeneous graphs.
RGCNs \cite{schlichtkrull_modeling_2018} and RGATs \cite{busbridge_relational_2019} model heterogeneous graphs by using a weight matrix for each relation. 
HAN \cite{wang_heterogeneous_2019} proposes a hierarchical attention mechanism to learn the weights of nodes and meta-paths from node-level and semantic-level attention, respectively. 
HetGNN \cite{zhang_heterogeneous_2019} adopts different Recurrent Neural Networks (RNNs) for different types of nodes to integrate multi-modal features.
Howerer, due to the large number of relations in KGs, the training complexity is high when applying them to model KGs.
More recently, HGT \cite{hu_heterogeneous_2020} and RHGT \cite{mei_relation-aware_2022} try to model the heterogeneity by heterogeneous graph transformers.
But they are not designed to capture neighborhood similarity, so it is difficult to directly apply to EA tasks.
Therefore, an improved heterogeneous graph transformer is designed to consider the heterogeneity of KGs, thereby obtaining high-quality entity embeddings for EA tasks.

\section{Proposed Framework}
\label{sec3}

\subsection{Problem Definition}

To increase the neighborhood semantics of entities, we introduce a meta path-based similarity framework for EA. 
The classic meta-path paradigm is defined as a sequence of relations between objects, so we define the new compound relation between two entities as a relation path in this paper.
For example, suppose $(e^1_h, e^2_k)$ is an aligned entity, where superscripts denote different KGs.
There is a path relation $(r^1_f,r^1_g)$ near $e^1_h$, because the following relation exists: $e^1_h \overset{r^1_f}{\rightarrow} e^1_i \overset{r^1_g}{\rightarrow} e^1_t$, but there may not be a similar path near $e^2_k$.
Therefore, not all paths in the neighborhood of entities are reliable for EA learning. 
In other words, we should only keep partially reliable paths to learn the neighborhood similarity of aligned entities.

\paragraph{Definition 1} (Reliable Path Set).
In this paper, we use $P \!=\! \{ p_1,..., p_i, ..., $ $p_N \}$ represents reliable path set, where each $p_i \!=\! (r_f,r_g)$ is effective for EA learning. 
“Reliable path” here refer to path that facilitate EA learning, rather than the meaningful path. 
We believe the paths that frequently occur in the neighborhoods of pre-aligned entities can be considered reliable.
In this paper, we only consider paths based on two-hop relations, and the study of a wider range of path structures will be left to future work.

\paragraph{Definition 2} (KG with Reliable Path Set).
We define KG as $G \!=\! (E,R,$ $T_{rel}, P, T_{path})$, where $E$ is entity set, $R$ is relation set, $T_{rel} \!\subseteq\! E \!\times\! R \!\times\! E$ is relation triple set, $P$ is reliable path set, and $T_{path} \!=\! \{ \langle e_h,p_k,e_t \rangle | p_k \!=\! (r_f, r_g) \!\in\! P, \langle e_h,r_f,e_a \rangle \!\in\! T_{rel}, \langle e_a,r_g,e_t \rangle \!\in\! T_{rel}\}$ is path triple set.

\paragraph{Definition 3} (Entity Alignment).
${G}^{1} \!=\! (E^1,R^{1},T^{1}_{rel},P^{1},$ $T^{1}_{path})$ and ${G}^{2} \!=\! (E^2, R^{2},T^{2}_{rel},P^{2},T^{2}_{path})$ are two KGs to be aligned.
Let $\mathbb{L} \!=\! \{ (e_i^{1}, e_j^{2}) | e_i^{1} \!\in\! E^{1}, e_j^{2} \!\in\! E^{2},e_i^{1} \!\equiv\! e_j^{2}\}$ be the pre-aligned entity set, where $\equiv$ refers to the same real-world object.
Entity Alignment tasks aim to find the remaining aligned entities between two KGs.

Formally, we use bold letter for embedding vector.
For example, $\mathbf{E}^1$ represents the embedding matrix of entities in ${G}^{1}$, and $\mathbf{e}^1_i$ represents the $i$-th row of $\mathbf{E}^1$.
In addition, the entity name is the most common text used to identify a entity, which can be used to effectively capture the semantic similarity of aligned entities.
Therefore, we apply pre-trained word embeddings to generate initial representations of entities, ${\mathbf {E}}^{n}$, and use them as the input of our framework.

\begin{figure*}[t]
	\centering
	\includegraphics[width=16cm]{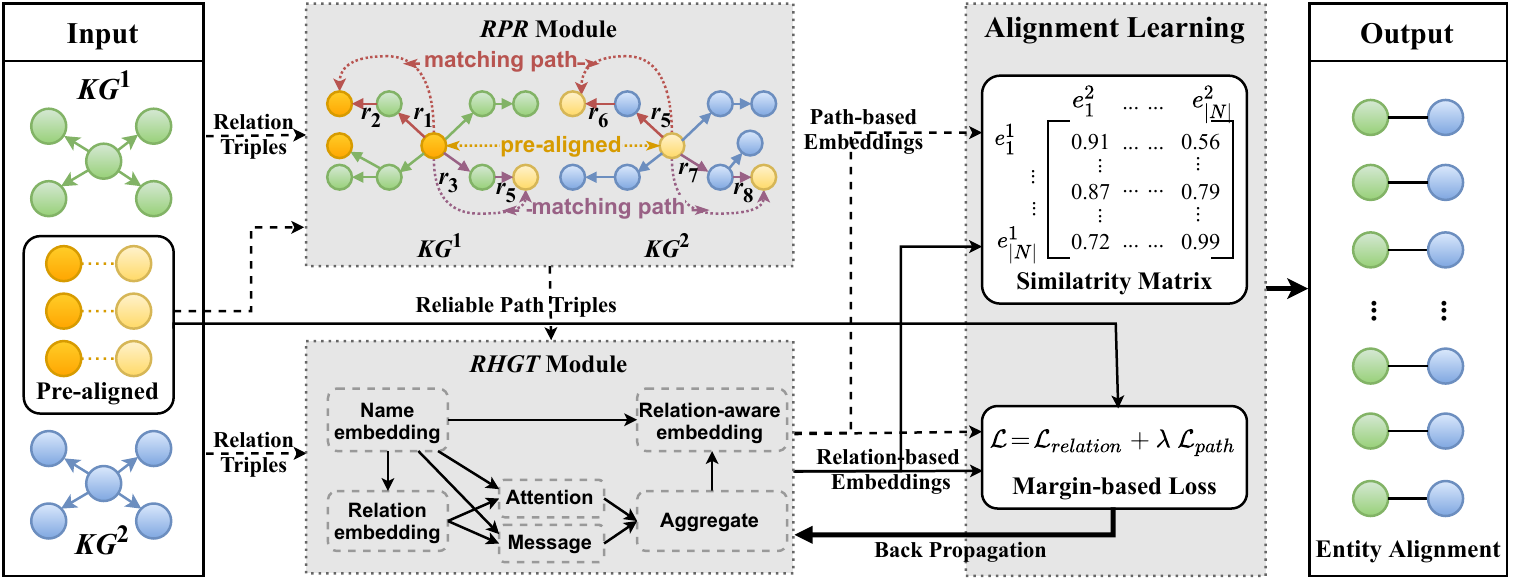}
	\caption{The overall architecture of \textit{RPR-RHGT}.}
	\label{fig2}
\end{figure*}
\subsection{Overview Framework of \textit{RPR-RHGT}}
In this section, we introduce our proposed framework \textit{RPR-RHGT}, a novel robust EA framework based on a reliable path reasoning algorithm and a relation-aware heterogeneous graph transformer.
Specifically, \textit{RPR-RHGT} is mainly composed of three modules, as shown in Figure \ref{fig2}:
(1) \textbf{Reliable Path Reasoning (\textit{RPR}).} 
A reliable path reasoning algorithm is developed to infer the reliable relation paths and form path structures of two KGs.
(2) \textbf{Relation-Aware Heterogeneous Graph Transformer (\textit{RHGT}).} 
We design the \textit{RHGT} to capture the features of specific patterns of relations and paths with fewer parameters, 
which contain the heterogeneous neighborhood features of aligned entities in relation and path structures.
(3) \textbf{Alignment Learning.}
This module computes the loss function and similarity matrices of path-based and relation-based entity embeddings, and evaluates the probabilities of EA.

\subsection{Reliable Path Reasoning (\textit{RPR})}

\begin{figure}[t]
	\centering
	\includegraphics[width=8.5cm]{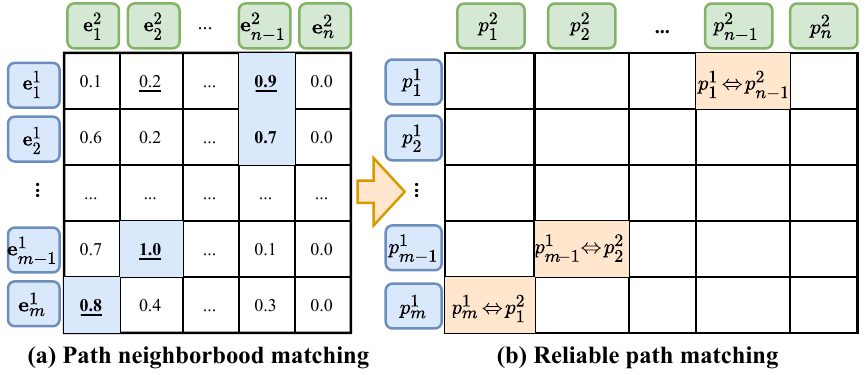}
	\caption{Illustration of the path neighborhood matching between the pre-aligned pair $(e^1_a, e^2_a) \in \mathbb{L}$.}
	\label{fig3}
\end{figure}

As discussed in Section \ref{sec1}, not all relation paths are reliable for EA learning.
It is known that each KG is constructed according to relatively stable data sources and construction rules.
Our key insight is the path with a high number of matches between the neighborhoods of pre-aligned entities (small range) can be regarded as reliable, which can be used to match judgments of other entities (large range).
We first establish the path neighborhood matching between each pre-aligned pair (see Figure \ref{fig3}(a)), derive the matching paths (see Figure \ref{fig3}(b)), finally select those paths with high numbers of matches to form a reliable path set $P$.

Specifically, for a given pair $(e^1_a, e^2_a) \in \mathbb{L}$, the similarity matrix $S$ denotes the similarities between path neighborhoods $PN(e^1_a)$ and $PN(e^2_a)$, where $PN(\cdot)$ indicates path neighborhood of a entity.
Firstly, the entities with maximum similarities in each row of $S$ are selected as the matching neighbors.
As shown in Figure \ref{fig3}(a), the matching result of $e^1_1$ (one neighbor of $e^1_a$) is $e^2_{n-1}$, because their similarity is the largest in first row.
However, there may be multiple neighbors of $e^1_a$ that match the same neighbor of $e^2_a$, such as $e^1_1$ and $e^1_2$ match with $e^2_{n-1}$ simultaneously. 
Therefore, the neighbor matching requires some one-to-one constraints:
1) the similarity of matching neighbor must reach a certain threshold: 
$MN(S) =\{e^2_{s^{max}_{k}}| s^{max}_{k} > \tau^{sim} \}$;
2) sort the similarity values that satisfy the threshold from high to low, and then perform one-to-one matching: $Match_{1:1}(MN(S)) = [(e^{1}_1,e^{2}_i), (e^{1}_2,e^{2}_j), ...]$.
So as a result, $e^1_1$ is chosen to match $e^2_{n-1}$, because $0.9 > 0.7$.
Obviously, only some neighbors of $e^1_a$ may end up finding matching neighbors.

Secondly, for each $(e^1_i, e^2_j) \in Match_{1:1}(MN(S))$, we can deduce the path matching pair $(p^1_k, p^2_t)$ according to the following reasoning relationship, as shown in Figure \ref{fig3}(b):
\begin{align}
e_i^{1} \leftrightarrow e_j^{2} \ \Rightarrow \ 
(e^1_a, p^1_k, e^1_i) \leftrightarrow (e^2_a, p^2_t, e_j^2) \ \Rightarrow \ 
p^1_k \leftrightarrow p^2_t,
\end{align}where
$ \leftrightarrow $ indicates the matching relationship;
$(e^1_a, p_k^{1}, e^1_i)$ and $ (e^2_a, p^2_t, e_j^2)$ are the path triples.

The last step is to count the matching number of each matching path, $counter(p^1_k \leftrightarrow p^2_t)$, and select those paths with high numbers of matches to form reliable path set $P$:
\begin{small}
	\begin{align}
	P = \{ (p^1_k, p^2_t) |counter(p^1_k \leftrightarrow p^2_t) > \tau^{path} \},
	\end{align} 
\end{small}where
$\tau^{path}$ is set according to the specific dataset.
Algorithm~\ref{alg1} gives the procedure of our \textit{RPR} algorithm.

\begin{algorithm}[h]
	\caption{Procedure of \textit{RPR} Algorithm.}
	\label{alg1}
	\textbf{Input}: (1) $G =\!(E,R,T_{rel})$;
	(2) pre-aligned entities $\mathbb{L}$;
	(3) entity name embeddings $\mathbf{E}^{n}$.\\
	\textbf{Output}: reliable path set $P$, path triple set $T_{path}$.
	
	\begin{algorithmic}[1]
		\STATE Set $P_{all} \leftarrow \emptyset $;
		\FOR {$ (e^1_a, e^2_a) \in \mathbb{L} $}
		\STATE Compute matching neighbors of path structures $Match_{1:1}\!(MN(S))$ between $PN(e^1_a)$ and $PN(e^2_a)$;
		
		\FOR {$ (e^1_i, e^2_j) \in Match_{1:1}(MN(S))$}
		\STATE Deduce the path matching pair $(p^1_k, p^2_t)$ using Eq.(1);
		\STATE $P_{all} \leftarrow P_{all} \cup {(p^1_k, p^2_t)}$;
		\ENDFOR
		\ENDFOR
		\STATE Generate the reliable path set $P$ using Eq. (2);
		\STATE Generate the path triple set $T_{path}$ using \textbf{Definition 2};
		\STATE \textbf{Return} $P$ and $T_{path}$;
	\end{algorithmic}
\end{algorithm}

\subsection{Relation-Aware Heterogeneous Graph Transformer (\textit{RHGT})}
The process of Graph Transformer \cite{yun_graph_2019} aggregating all neighborhood features of node $h$ can be briefly expressed as:
\begin{small} 
	\begin{align}
	\label{align1}
	\mathbf{e}^{(l)}_{h} \leftarrow 
	\underset{\forall t\in N(h)}{\mathbf{Aggregate}}\ (\mathbf{Attention}(h, t) \cdot \mathbf{Message}(h, t)),
	\end{align}
\end{small}where
\textbf{Attention} is to estimate the importance of each neighborhood node;
\textbf{Message} is to extract the feature of each neighborhood node;
and \textbf{Aggregate} aggregates the neighborhood message through attention weights.

\begin{figure}[t]
	\centering
	\includegraphics[width=8.5cm]{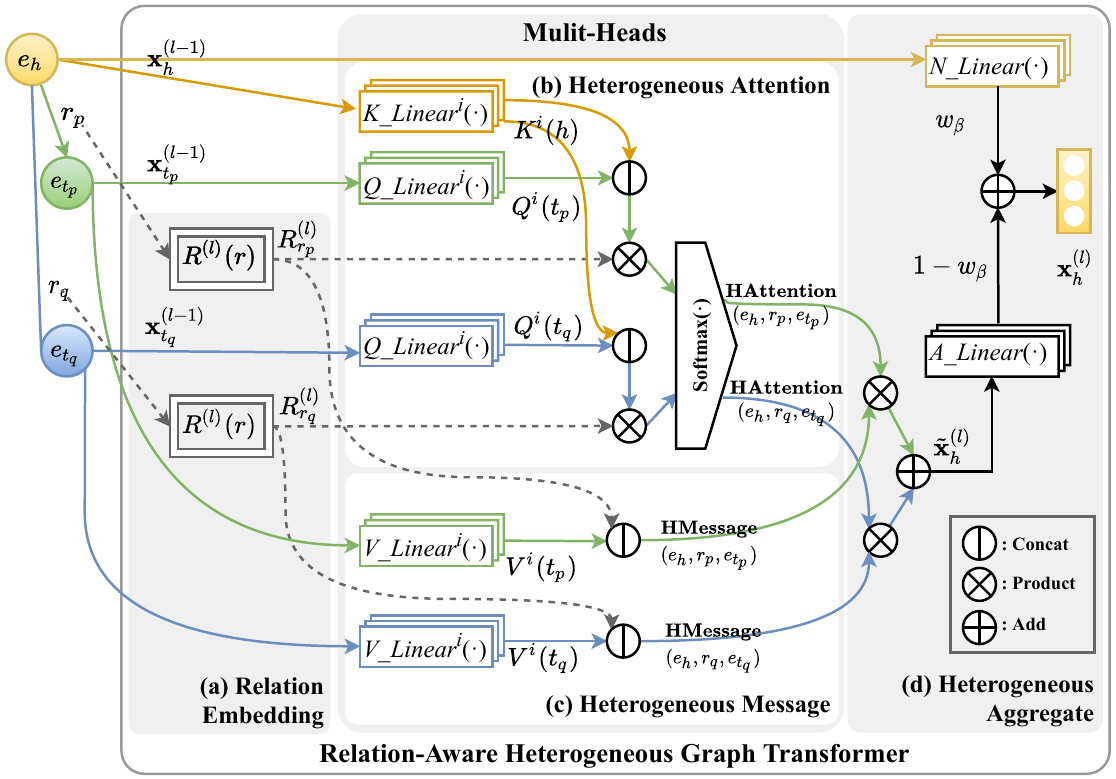}
	\caption{The overall architecture of \textit{RHGT}.}
	\label{fig4}
\end{figure}

As shown in Eq.(\ref{align1}), Graph Transformer does not consider the edge features.
Inspired by \cite{hu_heterogeneous_2020}, we design a relation-aware heterogeneous graph transformer (\textit{RHGT}), which enables our model to distinguish the heterogeneity features of relations and paths, to better obtain the neighborhood similarity of aligned entities.
Let $\mathbf{E}^{(l)}$ denote the output of $(l)$-th layer of \textit{RHGT}, which is also the input of the $(l+1)$-th layer.
Initially, $\mathbf{E}^{(0)} = {\mathbf {E}}^{n}$. 
When the input of \textit{RHGT} is the relation triples, the output is relation-based embeddings, and when the input is the path triples, the output is path-based embeddings.
As shown in Figure \ref{fig4}, \textit{RHGT} is mainly composed of four layers.

\paragraph{(a) Relation Embedding.}
Considering that the head entities and tail entities associated with aligned relations or aligned paths have certain similarities, we generate relation features by aggregating the features of associated entities.
Specifically, the embedding of $r$ is approximated by averaging the embeddings of its associated head entities $H_r$ and associated tail entities $T_r$ as: 
\begin{small}
	\begin{align}
	\label{align2}
	R^{l}(r) =\sigma \left[
	\frac{\sum_{e_i \in H_{r}} \mathbf{b_h} \mathbf{e}^{(l-1)}_{i}}{\left|H_{r}\right|}
	\| 
	\frac{\sum_{e_j \in T_{r}} \mathbf{b_t} \mathbf{e}^{(l-1)}_{j}}{\left|T_{r}\right|}
	\right],
	\end{align}
\end{small}where
$|\cdot|$ indicates the size of collection;
$\mathbf{b_h}, \mathbf{b_t}$ are the attention vectors;
$\| $ denotes concatenation and $\sigma $ is ReLU function.

\paragraph{(b) Heterogeneous Attention.}
Inspired by \cite{hu_heterogeneous_2020}, we map the entity $h$ into a key vector $K^{i}(h)$ and its neighborhood entity $t$ into a query vector $Q^{i}(t)$. 
The key difference from other methods is that instead of directly using the dot product of key and query vector as attention, we use the dot product between their concatenated result and $R^{l}(r)$.
$R^{l}(r)$ comes from the feature aggregation of the associated head and tail entities (see Eq.(\ref{align2})), so it will not deviate too far from the embeddings of its associated entities.
Moreover, $R^{l}(r)$ denotes heterogeneous features of edges, so neighbors associated with different edges contribute differently to the entity $h$.
Specifically, we compute the multi-head attention for each neighborhood relation $(h, r, t)$, as follows:
\begin{small}
	\begin{align}
	\label{align3}
	\begin{split}
	\mathbf{HAttention}(h, \!r, \!t) &\!= \! \underset{i \in [1,h_n]}{||} \! \ 
	\underset{\forall (r,t)\in \! RN\! (h)\! }{Softmax}(HATT_{head^{i}}(h, \!r, \!t)),
	\\
	HATT_{head^{i}}(h, \!r, \!t) &\!=\! \mathbf{a}^{T} \ ([K^{i}\!(h) || Q^{i}\!(t)] R^{(l)}\!(r)) / {\sqrt{ d/h_n}},
	\end{split}
	\end{align}
\end{small}where
$K^{i}(h) \!=\! {K\_Linear}^i (\mathbf{e}^{(l-1)}_{h})$;
$Q^{i}(t) \!=\! {Q\_Linear}^i$ $(\mathbf{e}^{(l-1)}_{t})$;
$RN(h)$ denotes the neighborhood of entity $h$;
$\mathbf{a} \in \mathbb{R}^{d/h_n \times 1} $ is the attention parameter;
$h_n$ is the number of attention heads and $d/h_n$ is the vector dimension per head.
Note that the Softmax process is to make the sum of attention vectors of all neighborhood entities equal to 1.

\paragraph{(c) Heterogeneous Message.}
Similarly, we hope to incorporate relations into the message passing process to distinguish the differences of different types of edges.
For any $(h, r, t) \in T$, its multi-head message is computed as follows:
\begin{small}
	\begin{align}
	\begin{split}
	\mathbf{HMessage}(h, r, t) &= \underset{i \in [1,h_n]}{||} \ HMSG_{Head^{i}}(h, r, t),
	\\
	HMSG_{Head^{i}}(h, r, t) &= [{V\_Linear}^i (\mathbf{e}^{(l-1)}_{t}) || R^{(l)}(r)].
	\end{split}
	\end{align}
\end{small}To 
get the $i$-th head message $HMSG_{Head^{i}}(h, r, t)$, we first apply a linear projection ${V\_Linear}^i$ to project the features of tail entity $t$, and then concatenate the features of $t$ and relation $r$.
The final heterogeneous message can be obtained by concatenating all $h_n$ message headers.

\paragraph{(d) Heterogeneous Aggregate.}
The final step is to aggregate heterogeneous multi-head attentions and messages of entities (see Figure \ref{fig4} (c)), thereby aggregating the information from neighbors with different feature to entity $h$.
The update vector $\tilde{\mathbf{e}}^{(l)}_{h}$ of $h$ can be obtained simply by averaging the corresponding messages from neighborhood entities with the attention coefficients as weights:
\begin{small}
	\begin{align}
	\tilde{\mathbf{e}}^{(l)}_{h} \!=\! 
	\underset{\forall (r,t) \in\! RN(h)}{\oplus} \mathbf{HAttention}(h, \!r,\! t) \!\cdot\! \mathbf{HMessage}(h, \!r, \!t) ,
	\end{align}
\end{small}where
$\oplus$ denotes the overlay operation.
To incorporate the name features and the features obtained by the multilayer neural network, the residual connection is used to generate the final updated embeddings as following: 
\begin{small}
	\begin{align}
	\mathbf{e}^{(l)}_{h} = w_{\beta } {A\_Linear}(\tilde{\mathbf{e}}^{(l)}_{h}) + (1- w_{\beta }) {N\_Linear}(\mathbf{e}^{(l-1)}_{h}),
	\end{align}
\end{small}where
$w_{\beta }$ is trainable weights,
$A\_Linear(\cdot),N\_Linear(\cdot)$ are linear projections.
Finally, we can generate relation-based embeddings $\mathbf{E}_{rel}$ and path-based embeddings $\mathbf{E}_{path}$ based on entire relation structure $T_{rel}$ and path structure $T_{path}$ respectively, and use them for end-to-end EA tasks.

\subsection{Alignment Learning}
After obtaining the final entity representations, we use Manhattan distance to measure the similarity of candidate entity pair. 
A smaller distance means a higher probability of entity alignment.
The following function is used to compute the similarity of candidate entity pair based on $\mathbf{E}_{rel}$ and $\mathbf{E}_{path}$:
\begin{small}
	\begin{align}
	d_f(e^1_i, e^1_j) = \| \mathbf{e}^1_{f,i} - \mathbf{e}^2_{f,j} \| _{L_1},
	\end{align}
\end{small}where 
$f \!=\! \{ rel, path \}$;
$L_1$ indicates the Manhattan distance.

To capture various aspects of the entities, previous methods usually concatenate the multi-source embeddings of entities and directly use them for the loss function.  
However, we argue that the contribution of relation-based and path-based embeddings to EA should be different, since these two structures of a entity may be quite diverse.
Therefore, instead of directly using concatenated embeddings, we assign different weights to the loss functions of the two embeddings, thereby distinguishing their different contributions during training.
In view of this, the following margin-based ranking loss function is used in model training, the goal of which is to keep the embedding distance of positive pair as small as possible and the embedding distance of negative pair as large as possible:
\begin{small}
	\begin{align}
	\begin{split}
	\mathcal {L} \!=\!
	\sum_{(p, q) \in \mathbb{L}, (p_{\prime}, q_{\prime}) \in \mathbb{L}_{rel}^{\prime}}
	[ d_{rel}(p, q) - d_{rel}(p^{\prime}, q^{\prime}) +{\gamma}_{1} ]_{+}
	\\ + \theta (
	\sum_{(p, q) \in \mathbb{L},(p_{\prime}, q_{\prime}) \in \mathbb{L}_{path}^{\prime}}
	[ d_{path}(p, q) - d_{path}(p^{\prime}, q^{\prime}) +{\gamma}_{2} ]_{+} \ ,
	\end{split}
	\end{align}
\end{small}where 
$[\cdot ]_{+} \!=\! max \{ 0,\cdot \}$;
$\mathbb{L}_{rel}^{\prime}$ and $\mathbb{L}_{path}^{\prime}$ represent the negative pair of relation-based and path-based embeddings, respectively;
${\gamma}_{1},{\gamma}_{2} \!>\! 0$ are the margin hyper-parameters for separating positive and negative pairs, respectively.

\section{Experiments}
\label{sec4}
In this section, we evaluate the performance of \textit{RPR-RHGT} on three widely used benchmark datasets.
The code is now available at \url{https://github.com/cwswork/RPR-RHGT}.

\subsection{Experiment Settings}

\begin{table}[t]
	\footnotesize
	\centering
	\setlength{\textwidth}{8.7cm}{
	\resizebox{\textwidth}{!}{
	\begin{tabular}{llr|rr|rr}
		\toprule
		\multirow{2}{*}{Datasets}	&\multirow{2}{*}{KGs}	&\multirow{2}{*}{Entities}  &\multicolumn{2}{c}{Relation}	&\multicolumn{2}{c}{Path}	\\
		&	&	&
		\multicolumn{1}{l}{Rels}	&\multicolumn{1}{l}{Triples}	&\multicolumn{1}{l}{Paths}	&\multicolumn{1}{l}{Triples}	\\
		
		\midrule
		\multirow{2}{*}{JA-EN(DBP)}	
		&JA		&65,744	&2,043	&164,373	&\multirow{2}{*}{139}	&283,311	\\
		&EN		&95,680	&2,096	&233,319	&						&266,759	\\
		\multirow{2}{*}{FR-EN(DBP)}	
		&FR		&66,858	&1,379	&192,191	&\multirow{2}{*}{172}	&559,984	\\
		&EN		&105,889	&2,209	&278,590	&					&505,443	\\
		\multirow{2}{*}{ZH-EN(DBP)}	
		&ZH		&66,469	&2,830	&153,929	&\multirow{2}{*}{140}	&166,991	\\
		&EN		&98,125	&2,317	&237,674	&						&436,418	\\
		\midrule
		\multirow{2}{*}{EN-DE(V1)}  
		&EN		&15,000	&215	&47,676	&\multirow{2}{*}{13}	&12,393	\\
		&DE		&15,000	&131	&50,419	&						&18,153	\\
		\multirow{2}{*}{EN-DE(V2)}  
		&EN		&15,000	&169	&84,867	&\multirow{2}{*}{38}	&58,517	\\
		&DE		&15,000	&96	 &92,632	&						&77,243	\\
		\multirow{2}{*}{EN-FR(V1)}  
		&EN		&15,000	&267	&47,334	&\multirow{2}{*}{46}	&51,349	\\
		&FR		&15,000	&210	&40,864	&						&50,504	\\
		\multirow{2}{*}{EN-FR(V2)}  
		&EN		&15,000	&193	&96,318	&\multirow{2}{*}{80}	&379,112	\\
		&FR		&15,000	&166	&80,112	&						&294,751	\\
		\midrule
		\multirow{2}{*}{DBP-WD}     
		&DBP	&100,000	&330	&463,294	&\multirow{2}{*}{460}	&1,834,831	\\
		&WD		&100,000	&220	&448,774	&						&2,709,929	\\
		\multirow{2}{*}{DBP-YG}     
		&YG		&100,000	&302	&428,952	&\multirow{2}{*}{115}	&1,148,939	\\
		&DBP	&100,000	&21		&502,563	&						&2,893,006	\\
		\bottomrule
	\end{tabular}
}}
	\caption{Statistics of datasets.}
	\label{tb1}
\end{table}

\paragraph{Datasets.}
Three experimental datasets contain cross-lingual datasets and mono-lingual dataset:
\textit{WK31-15K} \cite{sun_benchmarking_2020} is from multi-lingual DBpedia and used to evaluate model performance on sparse and dense datasets, where each subset contains two versions: V1 is sparse set obtained by using IDS algorithm, and V2 is twice as dense as V1.
\textit{DBP-15K} \cite{sun_cross-lingual_2017} is the most used dataset in the literature, and
is also from DBpedia.
\textit{DWY-100K} \cite{sun_bootstrapping_2018} contains two mono-lingual KGs, which serve as large-scale datasets to better evaluate the scalability of experimental models.
Table \ref{tb1} outlines the statistics of above datasets which also contains the numbers of paths and path triples generated by Algorithm~\ref{alg1}, to demonstrate the effect of the \textit{RPR} module.
Due to five-fold cross-validation used on \textit{WK31-15K} and \textit{DBP-15K}, the "Path.Paths" and "Path.Triplets" of these two datasets are the average statistic for the five training sets.

\paragraph{Metrics.}
Hits@$k$ is the proportion of correctly alignment ranked at the top-$k$ candidates;
MRR (Mean Meciprocal Rank) is the average of the reciprocal ranks.
Higher Hits@$k$ and MRR scores indicate better performance of EA.

\paragraph{Baselines.} 
For \textit{WK31-15K} and \textit{DBP-15K}, we compare \textit{RPR-RHGT} with eight previous state-of-the-art alignment models (mentioned in Section \ref{sec2}):
MTransE \cite{chen_multilingual_2017}, IPTransE \cite{zhu_iterative_2017}, 
JAPE \cite{sun_cross-lingual_2017}, BootEA \cite{sun_bootstrapping_2018},
AttrE \cite{trisedya_entity_2019}, RDGCN \cite{wu_relation-aware_2019}, 
NMN \cite{wu_neighborhood_2020}, RAGA \cite{zhu_raga_2021}.
Since only a few models are evaluated on \textit{DWY-100K}, we compare with the following models: 
MultiKE \cite{zhang_multi-view_2019}, RDGCN \cite{wu_relation-aware_2019}, NMN \cite{wu_neighborhood_2020}, COTSAE \cite{yang_cotsae_2020}.

\paragraph{Implementation Settings.}
For \textit{WK31-15K} and \textit{DBP-15K}, the proportion of train, validation and test is 2:1:7, the same as \cite{sun_benchmarking_2020}.
For \textit{DWY-100K}, we adopt the same train (30\%) / test (70\%) split as baselines. 
We use \textit{fastText} \footnote{\url{https://fasttext.cc/docs/en/crawl-vectors.html}} to generate entity name embeddings that are uniformly applied to baseline recurrence, including RDGCN, NMN, RAGA, MultiKE and COTSAE.
The embedding dimensions of 15K and 100K datasets are 300 and 200, respectively.
For all datasets, we use the same weight hyper-parameters: $\tau^{sim}=0.5, \tau^{path}=20, h_n \!=\! 4, \gamma_{1}\!=\! \gamma_{2}\!=\!10$, $\theta = 0.3$.

\subsection{Main Results}
\begin{table*}[t]
	\centering
	\setlength{\textwidth}{15cm}{
	\resizebox{\textwidth}{!}{
	\begin{tabular}{lrrrrrrrrrrrr}
		\toprule
		Datasets	
		&\multicolumn{3}{c}{EN-DE(V1)}		&\multicolumn{3}{c}{EN-DE(V2)}		
		&\multicolumn{3}{c}{EN-FR(V1)}		&\multicolumn{3}{c}{EN-FR(V2)}		\\
		Models	
		&\multicolumn{1}{c}{Hits@1} 	&\multicolumn{1}{c}{Hits@5}	&\multicolumn{1}{c}{MRR}
		&\multicolumn{1}{c}{Hits@1} 	&\multicolumn{1}{c}{Hits@5}	&\multicolumn{1}{c}{MRR}
		&\multicolumn{1}{c}{Hits@1} 	&\multicolumn{1}{c}{Hits@5}	&\multicolumn{1}{c}{MRR}
		&\multicolumn{1}{c}{Hits@1} 	&\multicolumn{1}{c}{Hits@5}	&\multicolumn{1}{c}{MRR}	\\
		
		\midrule
		*MTransE 
		&30.7	&51.8	&.407	
		&19.3	&35.2	&.274	
		&24.7	&46.7	&.351	
		&24.0	&43.6	&.336	\\
		
		\rowcolor{mygray}	
		*IPTransE 
		&35.0	&51.5	&.430	
		&47.6	&67.8	&.571	
		&16.9	&32.0	&.243	
		&23.6	&44.9	&.339	\\
		
		*JAPE 
		&28.8	&51.2	&.394	
		&16.7	&32.9	&.250	
		&26.2	&49.7	&.372	
		&29.2	&52.4	&.402	\\
		
		\rowcolor{mygray}	
		*BootEA 
		&67.5	&82.0	&.740	
		&83.3	&91.2	&.869	
		&50.7	&71.8	&.603	
		&66.0	&85.0	&.745	\\
		
		*AttrE 
		&51.7	&68.7	&.597	
		&65.0	&81.6	&.726	
		&48.1	&67.1	&.569	
		&53.5	&74.6	&.631	\\
		
		\rowcolor{mygray}	
		RDGCN 
		&81.98	&87.65	&.846	
		&81.61	&86.98	&.841	
		&80.53	&87.66	&.837	
		&87.12	&92.88	&.898	\\
		
		NMN 
		&85.57	&90.45	&.877	
		&\underline{85.18}	&\underline{89.57}	&\underline{.871}
		&\underline{85.12}	&\underline{90.74}	&\underline{.876}	
		&\underline{89.29}	&\underline{94.28}	&\underline{.915}	\\
		
		\rowcolor{mygray}	
		RAGA 
		&\underline{87.90}		&\underline{94.28}		&\underline{.908}	
		&81.34	&89.15	&.849	
		&82.71	&91.55	&.867	
		&88.95	&95.36	&.919	\\
		
		\midrule
		\textit{w/o.RPR}
		&90.26	&95.58	&.927	
		&92.08	&96.39	&.940	
		&88.31	&95.07	&.913	
		&93.60	&97.57	&.954	\\
		
		\rowcolor{mygray}	
		\textit{RPR-RHGT}	
		
		&\textbf{92.18}	&\textbf{96.32}	&\textbf{.940}	
		&\textbf{93.80}	&\textbf{97.20}	&\textbf{.953}	
		&\textbf{90.92}	&\textbf{95.54}	&\textbf{.930}	
		&\textbf{94.95}	&\textbf{98.00}	&\textbf{.963}	\\
		
		\midrule
		\textit{Improv. best}
		&4.28	&2.04	&.032	
		&8.62	&7.63	&.082	
		&5.80	&3.93	&.054	
		&5.66	&2.64	&.044	\\
		\bottomrule
	\end{tabular}
	}}
\setlength{\textwidth}{12cm}{
	\resizebox{\textwidth}{!}{
		\begin{tabular} {lrrrrrrrrr}
			\toprule
			Datasets	
			&\multicolumn{3}{c}{JA-EN(DBP)}		&\multicolumn{3}{c}{FR-EN(DBP)}		&\multicolumn{3}{c}{ZH-EN(DBP)}	\\
			Models	
			&\multicolumn{1}{c}{Hits@1}		&\multicolumn{1}{c}{Hits@5}		&\multicolumn{1}{c}{MRR}	
			&\multicolumn{1}{c}{Hits@1}		&\multicolumn{1}{c}{Hits@5}		&\multicolumn{1}{c}{MRR}	
			&\multicolumn{1}{c}{Hits@1}		&\multicolumn{1}{c}{Hits@5}		&\multicolumn{1}{c}{MRR}	\\
			\midrule
			MTransE 
			&20.41	&40.52	&.303		&19.74	&40.37	&.297		&20.89	&42.09	&.311	\\
			
			\rowcolor{mygray}	
			IPTransE 
			&27.92	&52.70	&.396		&31.22	&57.42	&.434		&17.34	&37.05	&.268	\\
			
			JAPE 
			&23.86	&44.50	&.340		&22.98	&45.22	&.336		&26.46	&50.30	&.378	\\
			
			\rowcolor{mygray}	
			BootEA 
			&52.71	&71.89	&.616		&57.61	&77.27	&.666		&55.45	&73.72	&.639	\\
			
			AttrE 
			&35.96	&60.31	&.475		&40.21	&66.09	&.522		&16.02	&33.29	&.250	\\
			
			\rowcolor{mygray}	
			RDGCN 
			&81.22	&87.98	&.844		&80.88	&88.08	&.842		&62.11	&73.88	&.676	\\
			
			NMN 
			&\underline{84.29}		&\underline{90.47}		&\underline{.870}	
			&83.46	&90.10	&.864	
			&65.16	&76.64	&.702	\\
			
			\rowcolor{mygray}	
			RAGA 
			&79.29	&89.12	&.838	
			&\underline{85.27}		&\underline{93.17}		&\underline{.889}	
			&\underline{68.72}		&\underline{82.55}		&\underline{.750}	\\
			
			
			\midrule
			\textit{w/o.RPR}	
			&87.43	&94.30	&.905		&87.69	&95.13	&.910		&66.85	&81.60	&.736	\\
			
			\rowcolor{mygray}	
			\textit{RPR-RHGT}	
			&\textbf{88.64}		&\textbf{94.30}		&\textbf{.912}	
			&\textbf{88.92}		&\textbf{95.59}		&\textbf{.919}	
			&\textbf{69.30}		&\textbf{82.66}		&\textbf{.754}	\\
			
			\midrule
			\textit{Improv. best}	
			&4.35	&3.83	&.042		&3.65	&2.42	&.030		&0.58	&0.11	&.004	\\
			\bottomrule
		\end{tabular}
}}

	\caption{	Overall performances of all models on \textit{WK31-15K} and \textit{DBP-15K}.
		“” marks the results obtained from OpenEA \protect \cite{sun_benchmarking_2020}.
		Other results are produced using their source code.	}
	\label{tb2}
\end{table*}

Tables \ref{tb2} and \ref{tb3} report all performances on three datasets.
The Hits@$k$ is in percentage (\%), while number in \textbf{bold} denotes the best results of all models and number in \underline{underline} denotes the best result of baselines.

\paragraph{Results on \textit{WK31-15K}.}
As shown in Table \ref{tb2}, \textit{RPR-RHGT} achieves the best performance on \textit{WK31-15K}, exceeding by 4.28\%$\sim $8.62\% on Hits@1.
By reducing the numbers of relations and triples, \textit{WK31-15K} challenges the ability of EA models to model sparse KGs. 
\textit{RPR-RHGT} achieves significant improvements over the baselines on both sparse KGs and dense KGs.
Besides, it is noteworthy that the improvements of \textit{RPR-RHGT} on Hits@1 are much higher than that on Hits@5, indicating that \textit{RPR-RHGT} can more accurately identify true entity among the top-5 indistinguishable alignment candidates.
This experiment shows \textit{RPR-RHGT} can compensate the neighborhood sparsity problem of some entities to a certain extent.

\paragraph{Results on \textit{DBP-15K}.}
From observing Table \ref{tb2}, the Hits@1 of \textit{RPR-RHGT} on \textit{DBP-15K} is higher than the best baselines by 4.35\%$\sim $0.58\%, which indicates that our model performs best on all \textit{DBP-15K}.
It is noteworthy that the performance of RAGA on \textit{ZH-EN(DBP)} is comparable to that of \textit{RPR-RHGT}.
We believe that \textit{ZH-EN(DBP)} has more mismatched paths as one of the reasons.
As shown in Table \ref{tb1}, \textit{ZH-EN(DBP)} has more relations than other datasets, but no more reliable paths obtained by \textit{RPR} algorithm.
Besides, NMN is one of the best performing baselines and effectively captures the cross-graph information and relation information of KG, while \textit{RPR-RHGT} still achieves good performance. 
Although the gap between \textit{RPR-RHGT} and RAGA is smaller, \textit{RPR-RHGT} has an advantage on \textit{DBP-15K}.

\begin{table}[h]
	\centering
	\setlength{\textwidth}{8.7cm}{
	\resizebox{\textwidth}{!}{
	\begin{tabular} {lrrrrrr}
		\toprule 
		Datasets		
		&\multicolumn{3}{c}{DBP-WD}		&\multicolumn{3}{c}{DBP-YG}\\
		Models	
		&\multicolumn{1}{c}{Hits@1}		&\multicolumn{1}{c}{Hits@10}		&\multicolumn{1}{c}{MRR}
		&\multicolumn{1}{c}{Hits@1}		&\multicolumn{1}{c}{Hits@10}		&\multicolumn{1}{c}{MRR}		\\
		\midrule
		MultiKE 
		&91.86	&96.26	&.935	
		&88.03	&95.32	&.906	\\
		
		\rowcolor{mygray}	
		RDGCN 
		&97.90	&99.10	&-	
		&94.70	&97.30	&-		\\
		
		NMN 
		&\underline{98.10}	&\underline{99.20}	&-		
		&\underline{96.00}	&\underline{98.20}	&-		\\
		
		\rowcolor{mygray}	
		COTSAE 
		&92.68	&97.86	&\underline{.945}	
		&94.39	&98.74	&\underline{.961}	\\
		
		\midrule
		\textit{w/o.RPR}	
		&99.11	&99.84	&.994	
		&96.30	&98.78	&.972		\\
		
		\rowcolor{mygray}	
		\textit{RPR-RHGT}							
		&\textbf{99.26}	&\textbf{99.86}	&\textbf{.995}	
		&\textbf{96.58}	&\textbf{98.86}	&\textbf{.974}		\\
		
		\midrule
		\textit{Improv. best}		
		&1.16	&0.66	&.050	
		&0.58	&0.12	&.013	\\
		
		\bottomrule
	\end{tabular}
	}}
	\caption{Overall performance of all models on \textit{DWY-100K}.
	All baseline performances are taken from their papers.}
	\label{tb3}
\end{table}

\paragraph{Results on \textit{DWY-100K}.}
As the largest dataset, \textit{DWY-100K} raises challenges to the time and space complexity of EA models.
As show in Table \ref{tb3}, although \textit{RPR-RHGT} does not rely on attribute structures, it still outperforms all baselines on \textit{DWY-100K}.
Since \textit{DWY-100K} is several times larger than other datasets, this experiment demonstrates that \textit{RPR-RHGT} has good scalability and superiority in larger real-world and monolingual KGs.


\subsection{Ablation Experiments}
\textit{w/o.RPR} is the \textit{RPR-RHGT} without \textit{RPR} module, the results of which are shown at the bottom of Tables \ref{tb2} and \ref{tb3}.
It can be observed that \textit{w/o.RPR} performs better than all baseline models on all datasets, except for \textit{ZH-EN(DBP)}, which confirms the effectiveness of \textit{RHGT} design. 
Besides, \textit{RPR-RHGT} achieves better performance than \textit{w/o.RPR} across all metrics and datasets.
This experiment confirms the assumption that the relational and path structure information of KGs can mutually reinforce each other.

\subsection{Further Analysis}

\begin{figure}[!t]
	\centering
	\includegraphics[width=8.5cm]{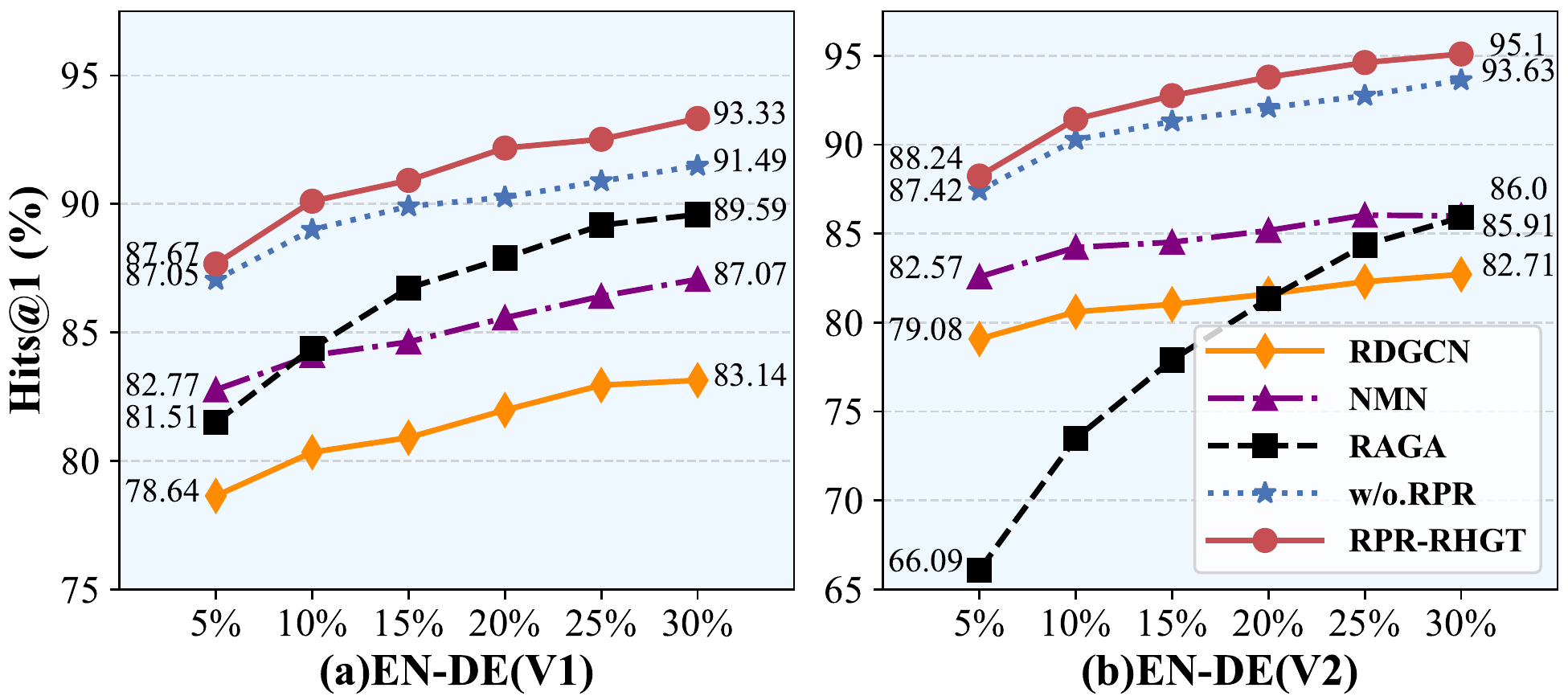}
	\caption{Performances with different ratios of pre-aligned entities.}
	\label{fig5} 
\end{figure}

\paragraph{Sensitivity to Ratios of Pre-Aligned Entities.}
To explore the impact of pre-aligned entities on EA model training, we implement a further evaluation based on different ratios of training set.
We take \textit{EN-DE(V1)} and \textit{EN-DE(V2)} as examples, and vary the ratio from 5\% to 30\%, while the validation dataset remains at 10\%.
RDGCN, NMN and RAGA are chosen as comparison models, all of which use name embeddings and perform best among baselines.
As shown in Figure \ref{fig5}, our two models maintain consistent performance, significantly outperforming the baselines on training sets for all scales.
This indicates that \textit{RPR-RHGT} can achieves satisfactory results based on fewer pre-aligned entities.

\begin{figure}[!t]
	\centering
	\includegraphics[width=8.5cm]{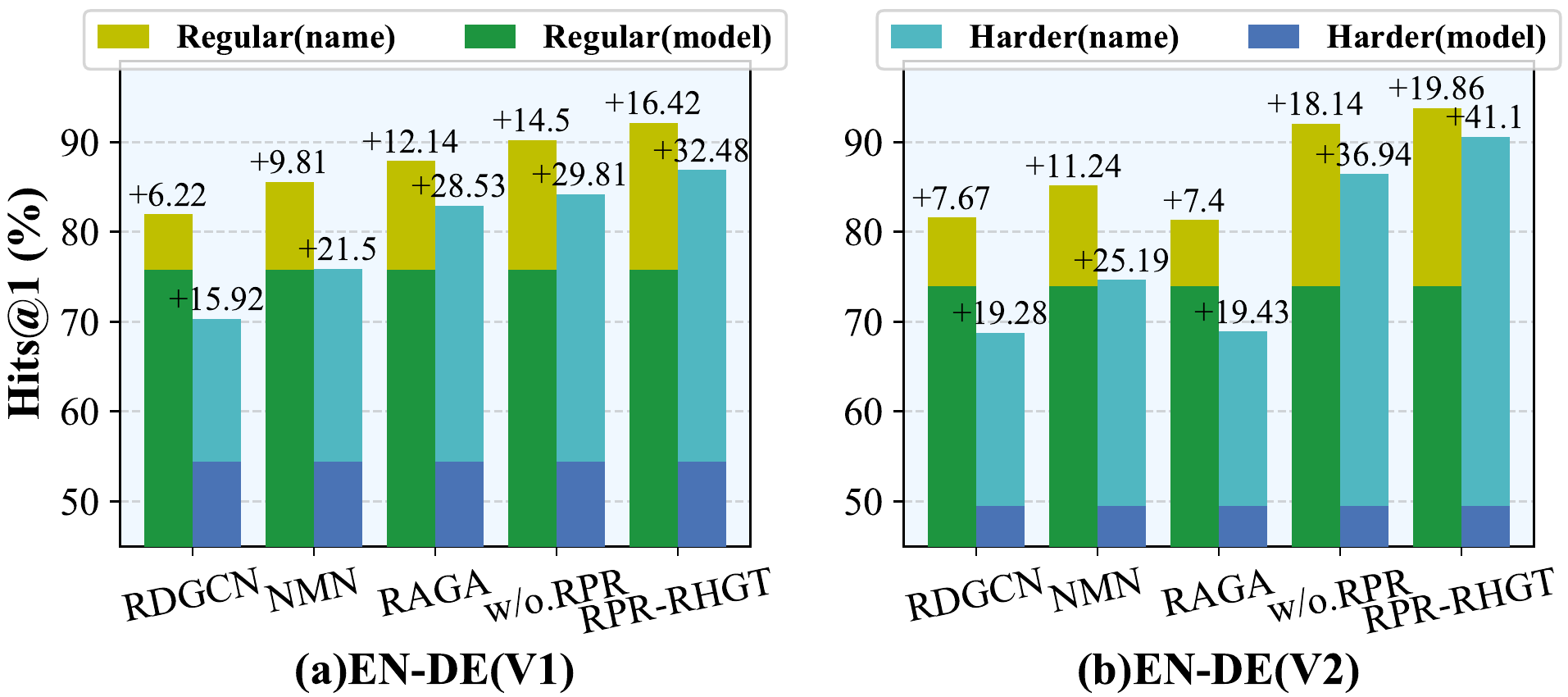}
	\caption{Hits@1 performances under regular and harder settings.}
	\label{fig6}
\end{figure}

\paragraph{Analysis on Harder Datasets.}
For a more objective evaluation of EA models, we take \textit{EN-DE(V1)} and \textit{EN-DE(V2)} as examples (called regular datasets), to construct two experimental datasets with relatively low similarities of entity names (called harder datasets).
Specifically, we first compute the name embedding similarities of aligned entity pairs and rank them (low to high), then pick the highest-ranked 50\% as the harder datasets, which are divided in the same way as above.
To compare the effects of name embeddings on the performances of regular and harder datasets, we also compute the alignment accuracy of entity embeddings based only on their name embeddings without training, i.e., \textit{Regular(name)} and \textit{Harder(name)}.
As shown in Figure \ref{fig6}, the performances of all models based on name embeddings drop on harder datasets.
However, comparing the performance on regular datasets, the performance of all models on harder dataset shows a more significant improvement over the performance of name embeddings.
In particular, \textit{RPR-RHGT} achieves up to 32.48\% and 41.1\% improvement over the name embeddings in Hits@1 on two harder datasets.
This result demonstrates the robustness of \textit{RPR-RHGT}, which can still promote effective EA on the datasets with less similar entity names.

\paragraph{Analysis on Training Time and Alignment Time.}
To evaluate the training and alignment efficiency of \textit{RPR-RHGT}, we compare the training time and alignment time of the following four models on \textit{EN-DE(V1)}.
The results running on a workstation with CPU (EPYC 3975WX +256G RAM) and GPU (RTX A4000 with 16G) are shown in Table \ref{tb4}, which shows large differences between different methods.
Although the training time of \textit{RPR-RHGT} is not optimal, its time complexity is competitive.
Overall, our model balances well between effectiveness and efficiency.

\begin{table}[]
	\centering
	\setlength{\textwidth}{8.7cm}{
	\resizebox{\textwidth}{!}{
	\begin{tabular}{lrrrrrr}
		\toprule 
		Model	
		&\multicolumn{2}{l}{\textbf{w/o.RPR}}	&\multicolumn{2}{l}{\textbf{RPR-RHGT}}	&\textbf{RAGA}	&\textbf{NMN} \\
		Take time(s)			&CPU		&GPU		&CPU		&GPU		&CPU		&CPU	\\
		\midrule
		Train of each epoch		&10.27	&0.27	&15.48	&0.40	&8.71	&29.74	\\
		Alignment of test set	&2.56	&1.92	&3.49	&2.04	&33.89	&101.49	\\
		\bottomrule
	\end{tabular}
	}}
	\caption{Comparison of training time and alignment time.}
	\label{tb4}
\end{table}
%

\section{Conclusions}
\label{sec5}

Traditional GNNs either donot consider the heterogeneous information of KGs, or cannot effectively extract heterogeneous information that is effective for EA tasks.
This paper proposes a new EA framework, \textit{RPR-RHGT}, which focuses on mining reliable path information and heterogeneous information, thereby making full use of KGs' own relation structures to improve alignment accuracy.
First, we develop a \textit{RPR} algorithm, which infers reliable paths from relation structures and only needs to be executed once. 
This algorithm is the first in the literature to successfully use unrestricted path information. 
Second, we improve a \textit{RHGT} model for modeling the heterogeneity of KGs, to better capture the heterogeneous neighborhood similarity of aligned entities.
Experimental results show \textit{RPR-RHGT} not only outperforms state-of-the-art models, but also achieves better performance in multiple ablation studies and analysis experiments.
In the future, we will continue to explore better ways to mine the heterogeneous information and path information of KGs for EA tasks.

\section*{Acknowledgments}
The works described in this paper are supported by The National Natural Science Foundation of China under Grant Nos. 61772210 and U1911201;
Guangdong Province Universities Pearl River Scholar Funded Scheme (2018);
The Project of Science and Technology in Guangzhou in China under Grant No. 202007040006.

\bibliographystyle{named} 
\bibliography{myalign}

\end{document}